%% file: main.tex
\documentclass[10pt,twocolumn,letterpaper]{article}

\usepackage{cvpr}              

\input{preamble}

\definecolor{cvprblue}{rgb}{0.21,0.49,0.74}
\usepackage[pagebackref,breaklinks,colorlinks,citecolor=cvprblue]{hyperref}

\usepackage{multirow}
\usepackage{amssymb}
\usepackage{amsmath}
\usepackage{xcolor}
\usepackage{caption}
\usepackage{graphicx}
\usepackage{float}

\def\confName{CVPR}
\def\confYear{2024}

\title{\LARGE \bf
TFNet: Exploiting Temporal Cues for Fast and Accurate \\ LiDAR Semantic Segmentation
}

\author{%
Rong Li{$^{1}$} ~ Shijie Li{$^{2}$\footnotemark[2]} ~ Xieyuanli Chen{$^{2}$} ~ Teli Ma{$^{1}$} ~ Juergen Gall{$^{2,4}$} ~ Junwei Liang{$^{1,3}$\footnotemark[2]}\\
\normalsize
$^{1}$\	HKUST(GZ), China ~~ $^{2}$\, University of Bonn, Germany  ~~ $^{3}$\, HKUST, China \\
\normalsize
$^{4}$\	Lamarr Institute for Machine Learning and Artificial Intelligence, Germany \\  
\normalsize
{\tt\small rli335@connect.hkust-gz.edu.cn} ~~
{\tt\small lsj@uni-bonn.de} ~~{\tt\small junweiliang@hkust-gz.edu.cn}
}

\begin{document}

\include{defines}
\maketitle

\renewcommand{\thefootnote}{\fnsymbol{footnote}}
\footnotetext[2]{~Corresponding authors.}

\input{sec/0_abstract}    
\input{sec/1_intro}

\input{sec/2_related}

\input{sec/3_method}

\input{sec/4_exps}

\input{sec/5_conclusion}

\paragraph{Acknowledgment}
This work was supported by the Meituan Academy of Robotics Shenzhen.
The views and conclusions contained herein are those of the authors and should not be interpreted as necessarily representing the official policies or endorsements, either expressed or implied, of Meituan. Shijie Li was supported by the Deutsche Forschungsgemeinschaft (DFG, German Research Foundation) GA1927/5-2 (FOR 2535 Anticipating Human Behavior).

{
    \small
    \bibliographystyle{ieeenat_fullname}
    \bibliography{bib}
}


\end{document}

%% file: preamble.tex
\usepackage[dvipsnames]{xcolor}


%% file: defines.tex
\definecolor{car}{rgb}{0.39215686, 0.58823529, 0.96078431}
\definecolor{bicycle}{rgb}{0.39215686, 0.90196078, 0.96078431}
\definecolor{motorcycle}{rgb}{0.11764706, 0.23529412, 0.58823529}
\definecolor{truck}{rgb}{0.31372549, 0.11764706, 0.70588235}
\definecolor{other-vehicle}{rgb}{0.39215686, 0.31372549, 0.98039216}
\definecolor{person}{rgb}{1.        , 0.11764706, 0.11764706}
\definecolor{bicyclist}{rgb}{1.        , 0.15686275, 0.78431373}
\definecolor{motorcyclist}{rgb}{0.58823529, 0.11764706, 0.35294118}
\definecolor{road}{rgb}{1.        , 0.        , 1.        }
\definecolor{parking}{rgb}{1.        , 0.58823529, 1.        }
\definecolor{sidewalk}{rgb}{0.29411765, 0.        , 0.29411765}
\definecolor{other-ground}{rgb}{0.68627451, 0.        , 0.29411765}
\definecolor{building}{rgb}{1.        , 0.78431373, 0.        }
\definecolor{fence}{rgb}{1.        , 0.47058824, 0.19607843}
\definecolor{vegetation}{rgb}{0.        , 0.68627451, 0.        }
\definecolor{trunk}{rgb}{0.52941176, 0.23529412, 0.        }
\definecolor{terrain}{rgb}{0.58823529, 0.94117647, 0.31372549}
\definecolor{pole}{rgb}{1.        , 0.94117647, 0.58823529}
\definecolor{traffic-sign}{rgb}{1.        , 0.        , 0.    }

\definecolor{nus00}{rgb}{0.9607843137254902, 0.9411764705882353, 1.0}
\definecolor{nus01}{rgb}{0.5647058823529412, 0.5019607843137255, 0.4392156862745098}
\definecolor{nus02}{rgb}{0.23529411764705882, 0.0784313725490196, 0.8627450980392157}
\definecolor{nus03}{rgb}{0.0, 0.27058823529411763, 1.0}
\definecolor{nus04}{rgb}{0.0, 0.6196078431372549, 1.0}
\definecolor{nus05}{rgb}{0.27450980392156865, 0.5882352941176471, 0.9137254901960784}
\definecolor{nus06}{rgb}{0.38823529411764707, 0.23921568627450981, 1.0}
\definecolor{nus07}{rgb}{0.5019607843137255, 0.0, 0.0}
\definecolor{nus08}{rgb}{0.30980392156862746, 0.30980392156862746, 0.1843137254901961}
\definecolor{nus09}{rgb}{0.0, 0.5490196078431373, 1.0}
\definecolor{nus10}{rgb}{0.2784313725490196, 0.38823529411764707, 1.0}
\definecolor{nus11}{rgb}{0.7490196078431373, 0.8117647058823529, 0.0}
\definecolor{nus12}{rgb}{0.29411764705882354, 0.0, 0.6862745098039216}
\definecolor{nus13}{rgb}{0.29411764705882354, 0.0, 0.29411764705882354}
\definecolor{nus14}{rgb}{0.23529411764705882, 0.7058823529411765, 0.4392156862745098}
\definecolor{nus15}{rgb}{0.5294117647058824, 0.7215686274509804, 0.8705882352941177}
\definecolor{nus16}{rgb}{0.0, 0.6862745098039216, 0.0}

\definecolor{poss01}{rgb}{ 0.11764705882352941, 0.11764705882352941, 1.0 }
\definecolor{poss02}{rgb}{ 0.7843137254901961, 0.1568627450980392, 1.0 }
\definecolor{poss03}{rgb}{ 0.9607843137254902, 0.5882352941176471, 0.39215686274509803 }
\definecolor{poss04}{rgb}{ 0.0, 0.23529411764705882, 0.5294117647058824 }
\definecolor{poss05}{rgb}{ 0.0, 0.6862745098039216, 0.0 }
\definecolor{poss06}{rgb}{ 0.0, 0.0, 1.0 }
\definecolor{poss07}{rgb}{ 0.5882352941176471, 0.9411764705882353, 1.0 }
\definecolor{poss08}{rgb}{ 0.0, 1.0, 0.49019607843137253 }
\definecolor{poss09}{rgb}{ 0.0, 0.7843137254901961, 1.0 }
\definecolor{poss10}{rgb}{ 1.0, 1.0, 0.19607843137254902 }
\definecolor{poss11}{rgb}{ 0.19607843137254902, 0.47058823529411764, 1.0 }
\definecolor{poss12}{rgb}{ 0.9607843137254902, 0.9019607843137255, 0.39215686274509803 }
\definecolor{poss13}{rgb}{ 0.5019607843137255, 0.5019607843137255, 0.5019607843137255 }

\makeatletter
\newcommand{\car@semkitfreq}{4.08}
\newcommand{\bicycle@semkitfreq}{0.02}
\newcommand{\motorcycle@semkitfreq}{0.04}
\newcommand{\truck@semkitfreq}{0.21}
\newcommand{\othervehicle@semkitfreq}{0.16}
\newcommand{\person@semkitfreq}{0.18}
\newcommand{\bicyclist@semkitfreq}{1.11e-6}
\newcommand{\motorcyclist@semkitfreq}{5.53e-9}
\newcommand{\road@semkitfreq}{19.87}  %
\newcommand{\parking@semkitfreq}{1.47}
\newcommand{\sidewalk@semkitfreq}{14.39}  %
\newcommand{\otherground@semkitfreq}{0.39}
\newcommand{\building@semkitfreq}{13.26}  %
\newcommand{\fence@semkitfreq}{7.23}
\newcommand{\vegetation@semkitfreq}{26.68}  %
\newcommand{\trunk@semkitfreq}{0.60}
\newcommand{\terrain@semkitfreq}{7.81} %
\newcommand{\pole@semkitfreq}{0.28}
\newcommand{\trafficsign@semkitfreq}{0.06}
\newcommand{\semkitfreq}[1]{{\csname #1@semkitfreq\endcsname}}

\newcommand{\bestimprove}[2]{\textbf{#1}$_\text{\textcolor{green!80!black}{~(+#2)}}$}
\newcommand{\secondimprove}[2]{\underline{#1}$_\text{\textcolor{teal}{~(+#2)}}$}
\newcommand{\thirdimprove}[2]{{#1}$_\text{\textcolor{gray}{~(+#2)}}$}
\newcommand{\noimprove}[2]{{#1}$_\text{\textcolor{red}{~(-#2)}}$}
\newcommand{\blankimprove}[2]{{#1}$_\text{\textcolor{teal}{~~~~~#2}}$}

\newcommand{\with}{\textcolor{green}{\ding{52}}}
\newcommand{\without}{\textcolor{red}{\ding{56}}}

%% file: sec/0_abstract.tex

\begin{abstract}

LiDAR semantic segmentation plays a crucial role in enabling autonomous driving and robots to understand their surroundings accurately and robustly. A multitude of methods exist within this domain, including point-based, range-image-based, polar-coordinate-based, and hybrid strategies. Among these, range-image-based techniques have gained widespread adoption in practical applications due to their efficiency. However, they face a significant challenge known as the ``many-to-one'' problem caused by the range image's limited horizontal and vertical angular resolution. As a result, around 20\% of the 3D points can be occluded. In this paper, we present TFNet, a range-image-based LiDAR semantic segmentation method that utilizes temporal information to address this issue. Specifically, we incorporate a temporal fusion layer to extract useful information from previous scans and integrate it with the current scan. We then design a max-voting-based post-processing technique to correct false predictions, particularly those caused by the ``many-to-one'' issue. We evaluated the approach on two benchmarks and demonstrated that the plug-in post-processing technique is generic and can be applied to various networks.    

\end{abstract}

%% file: sec/1_intro.tex
\section{INTRODUCTION}
\label{sec:intro}

\noindent LiDAR (light detection and ranging) semantic segmentation enables a precise and fine-grained understanding of the environment for robotics and autonomous driving applications~\cite{semantickitti, nuscenes, pmf}. %
There are four categories of methods:
point-based~\cite{pointnet,pointnet++,baafnet,spvcnn,kpconv,randla,latticenet}, range-image-based~\cite{cenet, fidnet, lenet,salsanext, rangenet, squeezeseg}, polar-based~\cite{polarnet} and hybrid methods~\cite{gfnet, cpgnet}.
Despite point-based methods achieving remarkable scores in metrics such as mean Intersection over Union (mIoU) and Accuracy, they tend to underperform in terms of computational efficiency. 
In contrast, the range-image-based methods are orders of magnitude more efficient than the other methods as substantiated by studies~\cite{analysis, rangeformer}.
This efficiency is further enhanced by the direct applicability of well-optimized Convolutional Neural Network (CNN) models, which strike a balance between speed and accuracy. 
Given the requirement of real-time performance and computational efficiency for ensuring safety in practical applications, the distinctive advantages of range-image-based methods make them a suitable choice for LiDAR semantic segmentation in real-world scenarios.

\begin{figure}
    \centering
    \includegraphics[width=0.95\columnwidth]{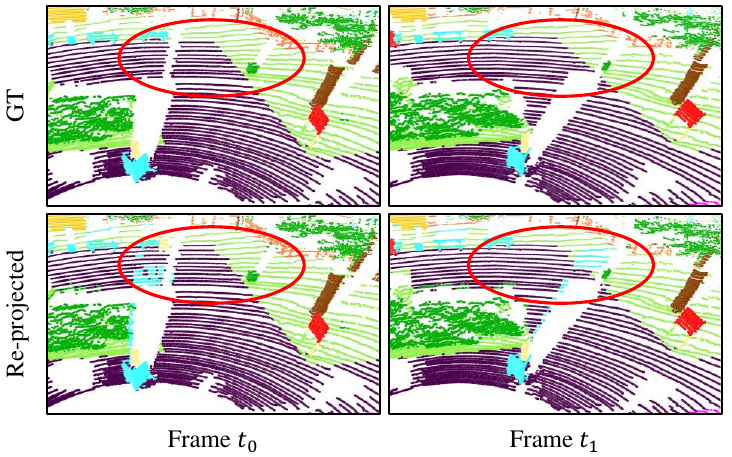}
    \caption{
    Range-image-based methods suffer from the ``many-to-one'' problem where multiple 3D points with the same angle are mapped to a single range pixel. Marked by the red circles of frame $t_0$, this can cause distant terrain points (purple) to receive erroneous predictions from nearby billboard points (blue) when the range image is re-projected to 3D.
    Furthermore, occluded points in frame $t_0$ become visible in $t_1$, offering an opportunity to refine the predictions.
    }
    \label{fig:top-figure}
\end{figure}

However, the range view representation suffers from a boundary-blurring effect~\cite{rangenet, fidnet}.
This problem exists mainly because of the limited horizontal and vertical angular resolution: more than one point will be projected to the same range image pixel when these points share the same vertical and horizontal angle. 
When multiple points share identical vertical and horizontal angles, they are projected onto the same pixel in the range image, giving rise to what is also referred to as the ``many-to-one'' problem~\cite{fidnet}.
Considering that the projection computes distant points first and near points later~\cite{rangenet}, the distant points will be occluded by the near points.
Hence, when converting the range image back into 3D coordinates, which is essential for range-image-based methods, the farther points receive the same label as the overlapping points that are closer. This leads to inaccuracies in the semantic understanding of the scene.

\cref{fig:top-figure} offers an illustration of this problem. Imagine that the LiDAR sensor is situated at the bottom-left of each range image. Close to the sensor, there is a billboard colored blue, and farther away is the terrain, displayed in purple. At time $t_0$, as marked by the red circle, even though the terrain and billboard are physically separate objects, some points on the terrain are incorrectly labeled as part of the billboard. This happens because these points, due to their similar angles relative to the LiDAR sensor, get projected onto the same pixel in the range image.
Upon the movement of the car and the consequent change in the sensor's field of view, we see a different scenario at time $t_1$: the previously mislabeled terrain points are now accurately classified. This improvement is attributed to the fact that their angular positions relative to the LiDAR sensor have changed, allowing them to avoid being hidden or masked by the billboard. This example illustrates how the dynamic movement affects LiDAR-based semantic segmentation and underscores the possibility of developing reliable and adaptable methods to tackle the ``many-to-one'' issue in this context.

We quantitatively assess the effects of this phenomenon on the SemanticKITTI dataset~\cite{semantickitti, behley2021ijrr}. Under standard conditions, where the range image dimensions are set to $64$ and $2048$ for height and width, respectively,   
it is observed that more than $20\%$ of the 3D points are occluded within the range image, i.e., more than one point is projected to the same pixel. As detailed in \cref{tab:post-process}, this results in a substantial degradation of the accuracy if it is not addressed by an additional post-processing step. 
Therefore, various post-processing approaches like k-NN~\cite{rangenet}, CRF~\cite{squeezeseg}, or NLA~\cite{fidnet} have been proposed. 
As an example, NLA~\cite{fidnet} resorts to assigning the label of the closest non-occluded point to occluded points. Nonetheless, this process necessitates checking each individual point for occlusion, which undermines the inherent efficiency of range-image-based methods. A detailed discussion about these methodologies can be found in Section \ref{sec:related}.

In this work, we propose to incorporate temporal information to address the ``many-to-one'' challenge for LiDAR semantic segmentation. This is inspired by human visual perception, where temporal information is crucial for understanding object motion and identifying occlusions. 
This is also observed in LiDAR semantic segmentation, where heavily occluded points can be captured from adjacent range image scans, as shown in~\cref{fig:top-figure}.
Based on this intuition, we exploit the temporal relations of features in the range map via cross-attention~\cite{attention, uniformer, mcmae}. 
As for the inference stage, we propose a max-voting-based post-processing scheme that effectively reuses the predictions of past frames. 
To this end, we transform the previous scans with predicted semantic class labels into the current ego car coordinate frame and then obtain the final segmentation by aggregating the predictions within the same voxel by max-voting.
In summary, we make the following three contributions:
\begin{itemize}
\item We quantitatively and qualitatively analyze and explain the ``many-to-one'' issues existing in range-image-based methods.
\item We propose TFNet, a range-based LiDAR semantic segmentation method. It utilizes a temporary cross-attention layer, which extracts informative features from previous LiDAR scans and combines them with current range features, to compensate for occluded objects.
\item We design a temporal-based post-processing method to solve the ``many-to-one'' mapping issue in range images. Compared with previous post-processing steps, our method achieves better performance, which is verified for various networks.
\item We evaluate the proposed method on two public benchmarks, namely SemanticKITTI~\cite{semantickitti} and SemanticPOSS~\cite{semanticposs}, where our method achieves a good trade-off between accuracy and inference time. 
\end{itemize}

%% file: sec/2_related.tex
\section{RELATED WORK}
\label{sec:related}

\noindent \textbf{LiDAR semantic segmentation.} 
The LiDAR sensor captures high-fidelity 3D structural information, which can be represented by various formats, i.e., points~\cite{pointnet, pointnet++, kpconv}, range view~\cite{squeezeseg,rangenet, cenet, fidnet, rangeformer}, voxels~\cite{minunet, cylindrical3d, spvcnn}, bird's eye view (BEV)~\cite{polarstream}, hybrid~\cite{gfnet, cpgnet} and multi-modal representations~\cite{2dpass, pmf, cmdfusion}. There are also some works~\cite{pmf, 2dpass} that fuse multi-sensor information. 
The point and voxel methods are prevailing, but their complexity is $\mathcal{O}(N\cdot d)$ where $N$ is in the order of $10^5$~\cite{analysis}. Thus, most approaches are not suitable for robotics or autonomous driving applications. The BEV method~\cite{polarstream} offers a more efficient choice with $\mathcal{O}(\frac{H\cdot W}{r^2}\cdot d)$ complexity, but the accuracy is subpar~\cite{rangeformer}. The multi-modal methods require additional resources to process the additional modalities. 
Among all representations, the range view reflects the LiDAR sampling process and it is much more efficient than other representations with $\mathcal{O}(\frac{H\cdot W}{r^2}\cdot d)$ complexity. We thus focus on the range-view as representation.

\noindent \textbf{Multi-frame LiDAR data processing.}
Multi-frame information plays a crucial role in LiDAR data processing. 
For example, MOS~\cite{mos} and MotionSeg3D~\cite{motionseg3d} generate residual images from multiple LiDAR frames to explore the sequential information and use it for segmenting moving and static objects. 
Motivated by these approaches, Meta-RangeSeg~\cite{meta-rangeseg} also uses residual range images for the task of semantic segmentation of LiDAR sequences. It employs a meta-kernel to extract the meta features from the residual images.
SeqOT~\cite{seqot} exploits sequential LiDAR frames using yaw-rotation-invariant OverlapNets~\cite{overlapnet, overlapnet2} and transformer networks~\cite{attention, overlaptrans} to generate a global descriptor for fast place recognition in an end-to-end manner.
In addition, SCPNet~\cite{scpnet} designs a knowledge distillation strategy between multi-frame LiDAR scans and a single-frame LiDAR scan for semantic scene completion.
Recently, Mars3D~\cite{mars3d} designed a plug-and-play motion-aware module for multi-scan 3D point clouds to classify semantic categories and motion states.
Seal~\cite{segment-any-point} proposes a temporal consistency loss to constrain the semantic prediction of super-points from multiple scans. Although the benefit of using multiple scans has been studied, these works address other tasks. 

\noindent \textbf{Post processing.}
Although range-view-based LiDAR segmentation methods are computationally efficient, they suffer from boundary blurriness or the ``many-to-one'' issue~\cite {squeezeseg, rangenet} as discussed in~\cref{sec:intro}.
To alleviate this issue, most works use a conditional random field (CRF)~\cite{squeezeseg} or k-NN~\cite{rangenet} to smooth the predicted labels.   
\cite{squeezeseg} implements the CRF as an end-to-end trainable recurrent neural network to refine the predictions according to the predictions of the neighbors within three iterations. It does not address occluded points explicitly.
k-NN~\cite{rangenet} infers the semantics of ambiguous points by jointly considering its k closest neighbours in terms of the absolute range distance. However, finding a balance between under and over-smoothing can be challenging, and it may not be able to handle severe occlusions.
Recently, NLA~\cite{fidnet} assigns the nearest point's prediction in a local patch to the occluded point. However, it is required to iterate over each point to verify occlusions.
In addition, RangeFormer~\cite{rangeformer} addresses this issue by creating sub-clouds from the entire point cloud and inferring labels for each subset. 
However, partitioning the cloud into sub-clouds ignores the global information. It can also not easily be applied to existing networks. 
Some methods~\cite{kprnet,motionseg3d,rangevit} propose additional refinement modules for the networks to refine the initial estimate, which increases the runtime.
In this work, we propose to tackle this issue by combining past predictions in an efficient max-voting manner. Our method complements existing approaches and can be applied to various networks.

%% file: sec/3_method.tex
\begin{figure*}[t!]
\centering
\includegraphics[width=1.0\linewidth]{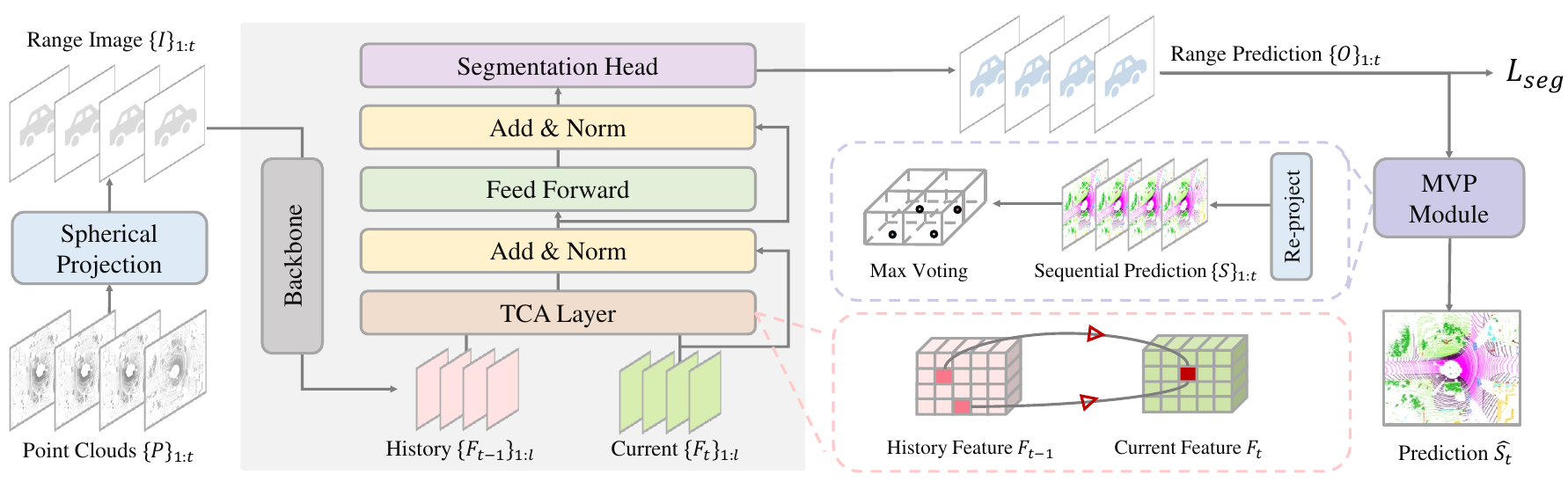}
    \caption{Architecture of TFNet. 
    For a point cloud $P_t$, TFNet projects it onto range images $I_t$. It then uses a segmentation backbone to extract multi-scale features $\{F_t\}_{1:l}$, a \textbf{T}emporal \textbf{C}ross-\textbf{A}ttention (\textbf{TCA}) layer to integrate past features $\{F_{t-1}\}_{1:l}$, and a segmentation head to predict range-image-based logits $O_t$. In inference, it refines the re-projected prediction ${S_t}$ 
    by aggregating the current and past temporal predictions $\{S\}_{1:t}$ by a \textbf{M}ax-\textbf{V}oting-based \textbf{P}ost-processing (\textbf{MVP}) strategy.
    } 
    \label{fig:arch}
\end{figure*}

\section{PROPOSED METHOD}

\subsection{Network Overview}

\noindent The overview of our proposed network is illustrated in~\cref{fig:arch}.
Our proposed network takes as input a point cloud $P$ comprising $N$ points represented by 3D coordinates $x, y, z$, and intensity $i$.
The point cloud is projected onto a range image $I$ of size $H \times W \times 5$ using a spherical projection technique employed in previous works~\cite{rangenet, squeezeseg}. 
Here, $H$ and $W$ represent the height and width of the image, and the last dimension includes coordinates $(x, y, z)$, range $r=\sqrt{x^2+y^2+z^2}$, and intensity $i$.
Next, we feed the range image into our backbone model to obtain multi-scale features $F$ with resolutions $\{1, 1/2, 1/4, 1/8\}$. 
We employ a \textbf{T}emporal \textbf{C}ross-\textbf{A}ttention (\textbf{TCA}) layer to integrate spatial features from the history frame. 
The aggregated features are then fed to the segmentation head, which predicts the range-image-based semantic segmentation logits $O$. 
For inference, we re-project the 2D semantic segmentation prediction to a 3D point-wise prediction $S$.
Subsequently, we propose a \textbf{M}ax-\textbf{V}oting-based \textbf{P}ost-processing (\textbf{MVP}) strategy to refine the current prediction $S_t$ by aggregating previous predictions.
We describe the key components of our network in the following sections.

\subsection{Temporal cross attention}

\noindent Although the range image suffers from the ``many-to-one'' issue, the occluded points can be captured from adjacent scans. 
This observation motivates us to incorporate sequential scans into both the training and inference stages.
First, we discuss how sequential data can be exploited during the training stage.

Inspired by the notable information extraction ability of the attention mechanism~\cite{attention} verified by various other works~\cite{segformer, uniformer, wang2023correlation, ma2023synchronize}, we use the cross-attention mechanism to model the temporal connection between the previous range feature $F_{t-1}$ and the current range feature $F_t$. 
The attended value is computed by:
\begin{equation}
    \mathbf{x}_{in} = {\rm Attention}({Q}, {K}, {V}) = {\rm Softmax}\left(\frac{{Q}\cdot  
 {K}^\mathsf{T}}{\sqrt{d_{f}}}\right){V}.
\label{eqn:mhsa}
\end{equation}
where $Q, K, V$ are obtained by $Q=Linear_q(F_t)$, $K=Linear_k(F_{t-1})$, $V= Linear_v(F_{t-1})$, and $d_f$ is the dimension of the range features.
We integrate a $3 \times 3$ convolution into the feed-forward module to encode positional information as in ~\cite{segformer} as well as a residual connection~\cite{resnet}.
The \textbf{feed-forward module} is defined as follows:
\begin{equation}
    {\mathbf{x}_{out} = {\text{MLP(GELU}(\text{Conv}_{\text{3} \times \text{3}}(\text{MLP}(\mathbf{x}_{in}))))+ \mathbf{x}_{in}}}.
    \label{eqn:ffn}
\end{equation}

The TCA module effectively exploits temporal dependencies in two ways. 
First, instead of using multiple stacked range features~\cite{mos}, our method extracts temporal information from the previous range features.
This not only reduces computational costs but also minimizes the influence of moving objects, which can introduce noise into the data.
Secondly, we only utilize the fusion module on the last feature level, which significantly decreases computation complexity. 
Previous works~\cite{uniformer, mcmae} have shown that the attention at shallower layers is not effective.

\subsection{Max-voting-based post-processing}\label{sec:mvp}

\begin{figure}[ht!]
\centering
\includegraphics[width=0.95\linewidth]{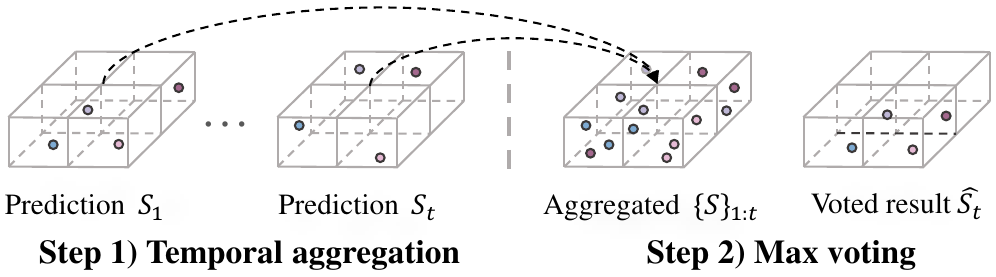}
    \caption{Illustration of the max voting post-processing strategy.}
    \label{fig:voting}
\end{figure}

\noindent While temporal cross attention exploits temporal information at the feature level, it does not resolve the ``many-to-one'' issue during the re-projection process of a range-image-based method, which causes occluded far points to inherit the predictions of near points.
We thus propose a max-voting-based post-processing (MVP) strategy, which is motivated by the observation that occluded points will be visible in the adjacent scans as shown in~\cref{fig:top-figure}.
As verified in~\cref{tab:scab-backbone}, MVP is generic and can be added to various networks.

\noindent{\textbf{Temporal scan alignment.}}
To initiate post-processing, it is essential to align a series of past LiDAR scans ($P_1, ..., P_t$) to the viewpoint (i.e., coordinate frame) of $P_t$. 
The alignment is accomplished by utilizing the estimated relative pose transformations $T_{j-1}^{j}$ between the scans $P_{j-1}$ and $P_j$. 
These transformation matrices can be acquired from an odometry estimation approach such as SuMa++~\cite{suma++}.
The relative transformations between the scans ($T_1^2, ..., T_{t-1}^{t}$) are represented by transformation matrices of $T_{j-1}^{j} \in \mathbb{R}^{4 \times 4}$. 
Further, we denote the $j^{th}$ scan transformed to the viewpoint of $P_t$ by
\begin{align}
P^{j \rightarrow t} & =\{{T}_j^t{p}_{i}\}_{{p}_i \in P_j} \quad
\text{with}~{T}_j^t = \prod_{k=j+1}^{t} {T}_{k-1}^{k}.
\label{eq:transformed_scan}
\end{align}

\noindent{\textbf{Sparse grid max voting.}}
After applying the transformations, we aggregate the aligned scans. We quantize the aggregated scans into a voxel grid with a fixed resolution $\delta$. 
In each grid, we use the max-voting strategy to use the most frequently predicted class label to represent the semantics of all points in the grid. 
We illustrate this process in~\cref{fig:voting} and evaluate the impact of the grid size in~\cref{fig:aba-grid-scans}. 
To save computation and memory, we store only the non-empty voxels. This sparse representation allows our method to handle large scenes.

\noindent{\textbf{Sliding window update.}}
We initialize a sliding window $W_{t-L+1:t}$ with the length of $L$ to store the scans and use a FIFO (First In First Out) strategy to update the points falling in each grid. 
When the LiDAR sensor obtains a new point cloud scan, we add it to this sliding window and remove the oldest scan. We do not use different weights across frames due to the uncertain occlusion problem. %

%% file: sec/4_exps.tex
\section{EXPERIMENTS}
\label{sec:exp}

\input{tables/semantickitti.tex}

\noindent {{\textbf{Datasets and evaluation metrics.}}} 
We evaluate our proposed method on SemanticKITTI~\cite{semantickitti} and SemanticPOSS~\cite{semanticposs}. 
SemanticKITTI~\cite{semantickitti} is a popular benchmark for LiDAR-based semantic segmentation in driving scenes. 
It contains 19,130 training frames, 4,071 validation frames, and 20,351 test frames. 
Each point in the dataset is provided with a semantic label of 19 classes for semantic segmentation. 
We also evaluate our dataset on the SemanticPOSS~\cite{semanticposs} dataset, which contains 2988 scenes for training and testing.
For evaluation, we follow previous works~\cite{cenet, rangeformer, fidnet, squeezeseg}, utilizing the class-wise Intersection over Union (IoU) and mean IoU (mIoU) metrics to evaluate and compare with others.

\noindent {{\textbf{Implementation details.}}} 
While we use CENet~\cite{cenet} as the main baseline method, our method demonstrates robust generalization across various backbones as shown in the following experiments.
We train the proposed method using the Stochastic Gradient Descent (SGD) optimizer and set the batch size to 8 and 4 for SemanticKITTI and SemanticPOSS, respectively. 
We follow the baseline method~\cite{cenet} to supervise the training with a weighted combination of cross-entropy, Lov\'asz softmax loss~\cite{lov-loss}, and boundary loss~\cite{boundary-loss}. The weights for the loss terms are set to $\beta_{1}=1.0$, $\beta_{2}=1.5$, $\beta_{3}=1.0$, respectively. 
All the models are trained on GeForce RTX 3090 GPUs. The inference latency is measured using a single GeForce RTX 3090 GPU. The backbone is trained from scratch on all the datasets. 

\subsection{Comparison with state of the art}
\label{sec:quanti}

\noindent{\textbf{Quantitative results on SemanticKITTI.}}
~\cref{tab:semantickitti} reports comparisons with representative models on the SemanticKITTI test set. Our method outperforms all range-image-based methods, including CNN-based architectures~\cite{cenet, fidnet, minet} and Transformer-based architectures~\cite{rangevit} in terms of mean IoU. 
CENet~\cite{cenet} uses test time augmentation to improve the performance. We do not use test time augmentation for a fair comparison with previous methods~\cite{rangenet, minet}.

~\cref{tab:semantickitti} presents a comprehensive comparison of the proposed TFNet method against several range-image-based LiDAR segmentation models on the SemanticKITTI test set. 
Specifically, TFNet excels in segmenting cars, bicycles, motorcycles, and pedestrians, showing significant improvements in IoU values over other methods. It registers particularly high IoU scores for bicycles (60.7\%), motorcycles (58.5\%), and persons (74.3\%).
Despite not always securing the top position in every class, TFNet consistently delivers strong results, especially in small and medium-sized object classes.
TFNet falls slightly behind in the pole and traffic-sign categories, where it records IoU scores lower than some methods like CENet~\cite{cenet} and KPRNet~\cite{kprnet}. Nevertheless, its ability to maintain balanced and above-average performance across most classes contributes to its overall leadership in mean-IoU.


\input{tables/semanticposs.tex}


\input{tables/postprocess}

\noindent{\textbf{Quantitative results on SemanticPOSS.}}
We present a quantitative evaluation of our TFNet method against several range-image-based LiDAR segmentation models on the SemanticPOSS test set~\cite{semanticposs} in~\cref{tab:poss}. 
Our method achieves the highest mean Intersection-over-Union (mIoU) among all listed methods, indicating overall better segmentation accuracy. Notably, TFNet excels in detecting smaller objects. It significantly surpasses CENet in segmenting traffic signs and poles, improving the IoU score by 6.9 percentage points and 6.3 percentage points, respectively. Furthermore, TFNet performs competitively in identifying cone/stone, achieving the second-best IoU score, closely following MINet's performance.
Moreover, TFNet ranks second in multiple categories such as rider, plants, fence, and bike, demonstrating its strong generalizability across diverse object classes.


\subsection{Ablation Analysis}
\label{sec:quanti}

\noindent{\textbf{Effect of the temporal post-processing.}}
\cref{tab:post-process} compares the proposed post-processing method with other post-processing approaches on the SemanticKITTI validation set. %
Using a CRF for post-processing has been used by SequeezeSegv2~\cite{squeezesegv2}.  
We train the network with CRF from scratch using the same training pipeline as our method.
The k-Nearest Neighbor (k-NN) method~\cite{rangenet} is the most popular post-processing method. It is widely used in Lite-HDseg~\cite{lite-hdseg}, SequeezeSegv3~\cite{squeezesegv3}, CENet~\cite{cenet}, SalsaNext~\cite{salsanext}, and MiNet~\cite{minet}. 
The Nearest Label Assignment (NLA) post-processing is used by FIDNet~\cite{fidnet}. It iterates over each point to check if a point is occluded or not. We use the source code from the corresponding methods.  
For the Point Refine module proposed in MotionSeg3D~\cite{motionseg3d}, we follow its implementation. We use SPVCNN~\cite{spvcnn} as the Point Refine module and use the features before the classification layer as the input to the Point Refine module. 
We then fine-tune the network with the Point Refine module in a second stage with a 0.001 learning rate for ten epochs. The results show the ``many-to-one'' issue harms the performance heavily. Without our proposed post-processing (`w/o MVP'), the mean IoU is $6.1$ lower. That CRF can actually decrease the mean IoU has also been shown in~\cite{rangenet}. While NLA and k-NN improve the results, the best mean IoU is achieved by our approach.      


\begin{table}[t!]
\centering
\renewcommand{\arraystretch}{1.3} 
\caption{Comparison with other temporal fusion methods.
}
\begin{tabular}{c|l}
\hline
Fusion Strategies        &       ~~mIoU~~    \\ \hline \hline
w/o TCA                &            66.9      \\ 
TMA module~\cite{tmanet} &     \thirdimprove{67.8}{0.9}             \\ 
Residual images~\cite{meta-rangeseg}        &     \noimprove{61.4}{5.5}   \\
Element-wise addition~\cite{mars3d}         &     \thirdimprove{67.6}{0.7}      \\
Channel concatenation~\cite{bevformerv2}    &     \secondimprove{68.0}{1.1}      \\
TCA module (ours)                     &      \bestimprove{68.1}{1.2}         \\ 
\hline
\end{tabular}
\label{tab:aba-fusion}
\end{table}
\noindent {\textbf{Effect of different fusion strategy}.} 
In~\cref{tab:aba-fusion}, we replace the proposed temporal fusion layer with other strategies. 
Mars3D~\cite{mars3d} adopts element-wise summation 
to aggregate temporal multi-scan point cloud embeddings and produce enhanced features.
The temporal memory attention (TMA) module~\cite{tmanet} validates its effectiveness on the video semantic segmentation task.
BEVFormer v2~\cite{bevformerv2} uses a feature warp and concatenation strategy to incorporate temporal information and shows its effectiveness on the LiDAR detection task. 
We follow its implementation, which concatenates previous BEV features with the current BEV feature along the channel dimension and employs residual blocks for dimensionality reduction.
We transform the scans to the same ego-car coordinates to implement the accurate alignment between temporal scans.
For the LiDAR semantic segmentation task, Meta-RangeSeg~\cite{meta-rangeseg} proposes to use three previous residual images as input and a meta-kernel module to incorporate temporal information. 
We follow its implementation and add to the five-channel input (x,y,z,r,i) three channels for the three residual images and a channel for the mask, which indicates whether the pixel is a projected 3D point or not.
The residual images are calculated by first transforming the point clouds of previous frames into the coordinates of the current frame and then calculating the absolute differences between the range values of the current scan and the transformed one with normalization.
A meta-kernel is followed to capture the spatial and temporal information. For a fair comparison, we keep the encoder and decoder of our architecture. 
We report the projection-based mIoU here because the loss function is applied directly to the range image.
All strategies are trained with the same setting and pipeline.
The results in \cref{tab:aba-fusion} show that our temporal fusion approach performs best.

\begin{table}[t!]
\centering
\renewcommand{\arraystretch}{1.5}  

\caption{Performance on other range-image-based methods.}
\resizebox{\columnwidth}{!}{
\begin{tabular}{c|ccc}
\hline
\multirow{2}{*}{Backbone} & \multicolumn{3}{c}{Post-processing}                                          \\ \cline{2-4} 
             & -  & k-NN~\cite{rangenet} & MVP (Ours) \\ 
\hline\hline
FIDNet~\cite{fidnet}  & ~55.4~  & \secondimprove{58.6}{3.2}    & \bestimprove{61.5}{6.1} \\
Meta-RangeSeg~\cite{meta-rangeseg}  & ~56.6~   & \secondimprove{60.3}{3.7}  & \bestimprove{63.1}{6.5}  \\
CENet~\cite{cenet}   & ~58.8~ & \secondimprove{62.6}{3.8}   & \bestimprove{64.7}{5.9}  \\
\hline
\end{tabular}
}
\label{tab:scab-backbone}
\end{table}

\subsection{Generalization Ability}

\cref{tab:scab-backbone} presents the effectiveness of the proposed max-voting-based post-processing (MVP) technique when integrated with three different range-image-based semantic segmentation methods, specifically FIDNet~\cite{fidnet}, Meta-RangeSeg~\cite{meta-rangeseg}, and CENet~\cite{cenet}. Unlike results reported in \cref{tab:semantickitti}, which reflect performances on the test set, this table displays the outcomes obtained on the validation set using publicly available pre-trained models with and without post-processing.
For each backbone model, the table compares three post-processing scenarios: no post-processing (denoted as `-'), application of the k-NN method from~\cite{rangenet}, and our proposed MVP. Each row shows the mean Intersection-over-Union (IoU) scores resulting from these treatments.

It is evident from the table that employing the MVP consistently leads to notable improvements over the baseline scores (without any post-processing) and often surpasses the performance of k-NN post-processing. For instance, MVP increases the IoU score of FIDNet by 6.1 points compared to its base result, demonstrating superior refinement capabilities. Similarly, the IoU scores of Meta-RangeSeg and CENet also witness considerable boosts with the use of MVP, affirming its broad applicability and positive impact on various range-image-based semantic segmentation models.


\subsection{Further Analysis}

\begin{figure}[t!]
\centering
\includegraphics[width=0.99\linewidth]{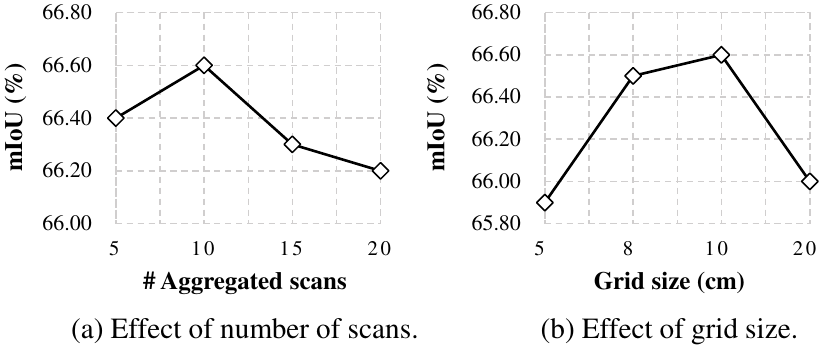}
    \caption{Effect of window size and grid size resolution.}
    \label{fig:aba-grid-scans}
\end{figure}

\noindent {{\textbf{Effect of frame numbers.}}} 
In~\cref{fig:aba-grid-scans}~(a), we delve into the effect of frame numbers, investigating the optimal length L of the sliding window used for temporal updates. This parameter determines the number of consecutive LiDAR frames that are combined to exploit temporal coherence in the scene as described in \cref{sec:mvp}. 
Our analysis reveals that setting L to 10 frames achieves a desirable balance between capturing sufficient temporal context and avoiding excessive computational load or memory requirements. This optimal choice also enables the model to effectively leverage temporal dependencies while maintaining real-time performance and reducing potential noise introduced by distant past or future frames.

\noindent{\textbf{Effect of grid size resolution.}}
As mentioned in \cref{sec:mvp}, we convert the accumulated LiDAR scans into a voxel grid format with a fixed resolution. It is crucial to select an appropriate resolution because the fundamental assumption is that all points enclosed within a voxel belong to the same semantic category. Overestimating the voxel size can undermine this assumption, whereas selecting a resolution that is too fine can introduce noise into the estimates due to the inclusion of small-scale variations.
To investigate the consequences of different voxel sizes, we perform an evaluation showcased in \cref{fig:aba-grid-scans}(b). The results clearly demonstrate that a voxel resolution of 0.10 meters yields the best semantic segmentation outcome. This finding underscores the significance of carefully tuning the grid size resolution to ensure that it neither oversimplifies nor overcomplicates the representation of the point cloud data, thereby preserving the integrity and accuracy of the semantic segmentation task.


\begin{figure*}[!htbp]
 \centering
\includegraphics[width=1.0 \linewidth]{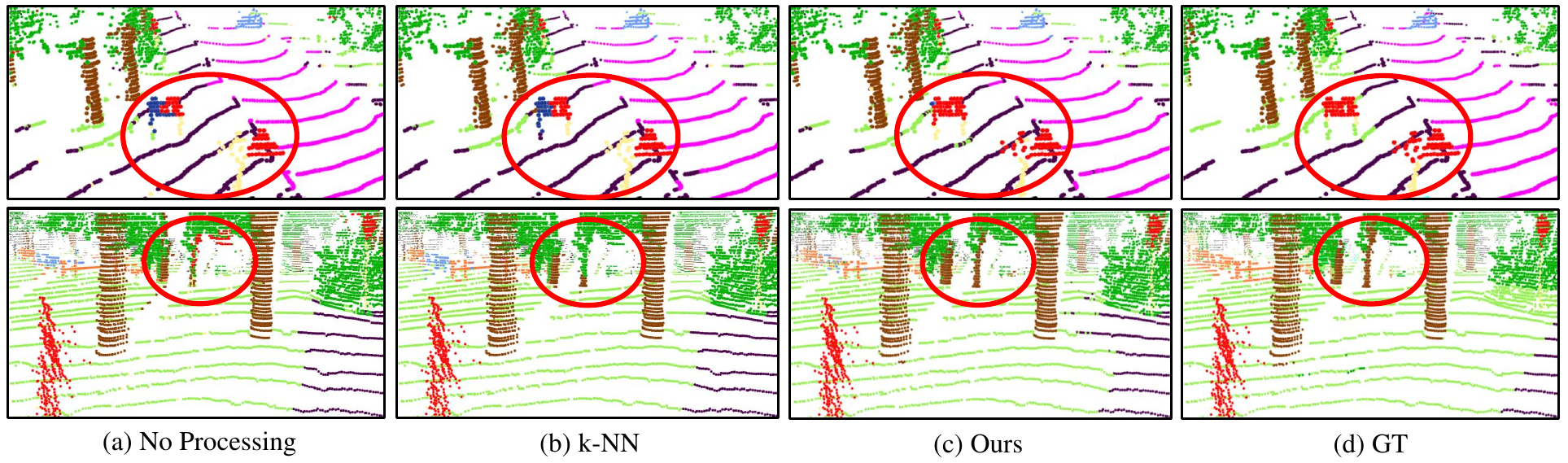}

\scriptsize
\textcolor{bicycle}{$\blacksquare$}~bicycle~
\textcolor{car}{$\blacksquare$}~car~
\textcolor{motorcycle}{$\blacksquare$}~motorcycle~
\textcolor{truck}{$\blacksquare$}~truck~
\textcolor{other-vehicle}{$\blacksquare$}~other vehicle~
\textcolor{person}{$\blacksquare$}~person~
\textcolor{bicyclist}{$\blacksquare$}~bicyclist~
\textcolor{motorcyclist}{$\blacksquare$}~motorcyclist~
\textcolor{road}{$\blacksquare$}~road~
\textcolor{parking}{$\blacksquare$}~parking~
\\
\scriptsize
\textcolor{sidewalk}{$\blacksquare$}~sidewalk~
\textcolor{other-ground}{$\blacksquare$}~other ground~
\textcolor{building}{$\blacksquare$}~building~
\textcolor{fence}{$\blacksquare$}~fence~
\textcolor{vegetation}{$\blacksquare$}~vegetation~
\textcolor{trunk}{$\blacksquare$}~trunk~
\textcolor{terrain}{$\blacksquare$}~terrain~
\textcolor{pole}{$\blacksquare$}~pole~
\textcolor{traffic-sign}{$\blacksquare$}~traffic sign			

\caption{Qualitative analysis of the post-processing scheme. (a) The ``many-to-one'' issue is evident without post-processing, e.g., the trunk is partially segmented as traffic sign and vegetation as they project onto the same range pixel (row 2).
(b) k-NN~\cite{rangenet} smooths the semantic labels locally, but it cannot resolve ambiguities by objects that are close or prediction errors. %
(c) Our method exploits temporal information to resolve false predictions (row 1) or ambiguities due to occlusions (row 2). 
Best viewed in color.}
\label{fig:comp-vis-post}
\end{figure*}


\begin{figure}[ht!]
\centering
\resizebox{1\linewidth}{!}{
\includegraphics[width=1.0\linewidth]{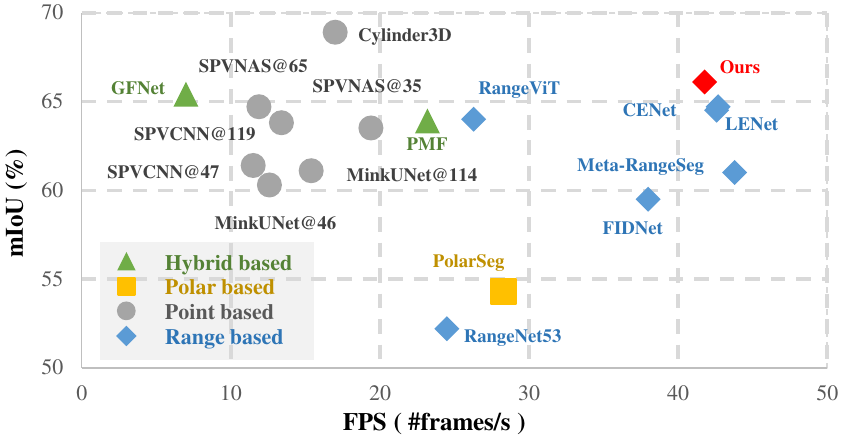}
}
\caption{mIoU vs.\ runtime on SemanticKITTI. 
Our method balances mIoU and inference time better than other state-of-the-art methods.
Best viewed in color.
}
\label{fig:fps-iou}
\end{figure}


\noindent{\textbf{Inference time comparison.}}
We visualize popular methods' inference time and mIoU in~\cref{fig:fps-iou}.
The results show that range-image-based methods are faster than point, polar, or hybrid methods.
We measured the inference time of all the methods on the same hardware with a GeForce RTX 3090 GPU for a fair comparison. 

\noindent{\textbf{Qualitative evaluation.}}
\label{sec:quali}
The ``many-to-one'' issue becomes apparent in the absence of any post-processing technique, as depicted in \cref{fig:comp-vis-post}(a). Here, we observe that points belonging to the tree trunk inadvertently adopt the predictions intended for nearby points from the traffic sign and vegetation classes. This occurs because these points, despite being distinct physical entities, are projected onto the same range pixel in the LiDAR data.
In \cref{fig:comp-vis-post}(b), we illustrate the performance of the commonly-employed k-NN method~\cite{rangenet}. While it does refine the initial predictions to some extent, it struggles to rectify false classifications when larger regions are occluded. This limitation highlights the inability of certain post-processing methods to handle complex scenarios where multiple points project to the same pixel.
On the contrary, our proposed method effectively tackles this problem, as shown in \cref{fig:comp-vis-post}(c). By incorporating temporal information across multiple scans, our approach consistently maintains the correct predictions for the tree trunk, even when the current scan is affected by the ``many-to-one" issue. This capability showcases the merit of introducing temporal context in the post-processing phase, as it allows our method to discern and rectify errors caused by occlusions and projection ambiguities in LiDAR data. Thus, our solution demonstrates improved robustness in handling the ``many-to-one" problem, illustrating the potential gains achieved by leveraging temporal coherence in LiDAR semantic segmentation.

%% file: tables/semantickitti.tex
\begin{table*}[ht!]
\centering
\large
\renewcommand{\arraystretch}{1.5} 
\caption{Comparison with other range-image-based LiDAR segmentation methods with resolution $(64,2048)$ on SemanticKITTI test set.}
\resizebox{\textwidth}{!}{
 \begin{tabular}{l| c| c | c c c c c c c c c c c c c c c c c c c} 
 \hline
  & \rotatebox{90}{mean-IoU}  & \rotatebox{90}{car}& \rotatebox{90}{bicycle}& \rotatebox{90}{motorcycle}& \rotatebox{90}{truck}& \rotatebox{90}{other-vehicle}& \rotatebox{90}{person}& \rotatebox{90}{bicyclist}& \rotatebox{90}{motorcyclist}& \rotatebox{90}{road}& \rotatebox{90}{parking}& \rotatebox{90}{sidewalk}& \rotatebox{90}{other-ground}& \rotatebox{90}{building}& \rotatebox{90}{fence}& \rotatebox{90}{vegetation}& \rotatebox{90}{trunk}& \rotatebox{90}{terrain}& \rotatebox{90}{pole}& \rotatebox{90}{traffic-sign} \\ 
 \hline\hline
    MINet~\cite{minet} & 55.2  & 90.1 & 41.8 & 34.0 & 29.9 & 23.6 & 51.4 & 52.4 & 25.0 & 90.5 & 59.0 & 72.6 & 25.8 & 85.6 & 52.3 & 81.1 & 58.1 & 66.1 & 49.0 & 59.9  \\
    
    FIDNet~\cite{fidnet} & 59.5  & 93.9 & 54.7 & 48.9 & 27.6 & 23.9 & 62.3 & 59.8 & 23.7 & 90.6 & 59.1 & 75.8 & 26.7 & 88.9 & 60.5 & 84.5 & 64.4 & 69.0 & 53.3 & 62.8\\
    
    Meta-RangeSeg~\cite{meta-rangeseg} & 61.0  & 93.9 & 50.1 & 43.8 & \underline{43.9} & 43.2 & 63.7 & 53.1 & 18.7 & 90.6 & 64.3 & 74.6 & 29.2 & 91.1 & 64.7 & 82.6 & 65.5 & 65.5 & 56.3 & 64.2\\
    
    KPRNet~\cite{kprnet} & 63.1  & \textbf{95.5} & 54.1 & 47.9 & 23.6 & 42.6 & 65.9 & 65.0 & 16.5 & \textbf{93.2} & \textbf{73.9} & \textbf{80.6} & 30.2 & \underline{91.7} & \underline{68.4} & \textbf{85.7} & 69.8 & \textbf{71.2} & 58.7 & 64.1 \\
    
    Lite-HDSeg~\cite{lite-hdseg} & 63.8 & 92.3& 40.0& \underline{55.4}& 37.7& 39.6& 59.2& \underline{71.6} & \textbf{54.1} & 93.0& 68.2& 78.3& 29.3& 91.5& 65.0& 78.2& 65.8& 65.1 & 59.5 & \underline{67.7}\\
    
    CENet~\cite{cenet} & \underline{64.7}  & 91.9 & 58.6 & 50.3 & 40.6 & 42.3 & \underline{68.9} & 65.9 & \underline{43.5} & 90.3 & 60.9 & 75.1 & \underline{31.5} & 91.0 & 66.2 & 84.5 & 69.7 & 70.0 & 61.5 & 67.6\\
    
    RangeViT~\cite{rangevit} & 64.0  & 95.4 & 55.8 & 43.5 & 29.8 & 42.1 & 63.9 & 58.2 & 38.1 & \underline{93.1} & 70.2 & \underline{80.0} & \textbf{32.5} & \textbf{92.0} & \textbf{69.0} & \underline{85.3} & \underline{70.6} & \underline{71.2} & \underline{60.8} & 64.7 \\

    LENet~\cite{lenet} & 64.5  & 93.9 & 57.0 & 51.3 & \textbf{44.3} & \underline{44.4} & 66.6 & 64.9 & 36.0 & 91.8 & 68.3 & 76.9 & 30.5 & 91.2 & 66.0 & 83.7 & 68.3 & 67.8 & 58.6 & 63.2\\
    
    TFNet~(Ours) & \textbf{66.1}  & \underline{94.3} & 60.7 & \textbf{58.5} & 38.4 & \textbf{48.4} & \textbf{74.3} & \textbf{72.2} & 35.5 & 90.6 & \underline{68.5} & 75.3 & 29.0 & 91.6 & 67.3 & 83.8 & \textbf{71.1} & 67.0 & \underline{60.8} & \textbf{68.7} \\

 \hline
\end{tabular}
}
\label{tab:semantickitti}
\end{table*}

%% file: tables/semanticposs.tex
\begin{table}[t!]
\centering
\renewcommand{\arraystretch}{1.5} 
\caption{Evaluation results on the SemanticPOSS test set.}
\label{tab:poss}

\resizebox{\columnwidth}{!}{
\begin{tabular}{c|c|c|c|c|c|c|c}
\hline
~ & 
\rotatebox{90}{Sq.Seg~\cite{squeezeseg}} &
\rotatebox{90}{Sq.SegV2~\cite{squeezesegv2}} &
\rotatebox{90}{RangeNet~\cite{rangenet}} &
\rotatebox{90}{MINet~\cite{minet}} &
\rotatebox{90}{FIDNet~\cite{fidnet}} &
\rotatebox{90}{CENet~\cite{cenet}} &
\rotatebox{90}{TFNet~(Ours)} \\ 
\hline\hline
person & 6.8 & 43.9 & 57.3 & 62.4 & 72.2 & \textbf{75.5} & \underline{72.4} \\
rider & 0.6 & 7.1 & 4.6 & 12.1 & \textbf{23.1} & \underline{22.0} & 20.5 \\
car & 6.7 & 47.9 & 35.0 & 63.8 & 72.7 & \underline{77.6} & \textbf{77.7} \\ 
truck & 4.0 & 18.4 & 14.1 & 22.3 & 23.0 & \textbf{25.3} & \underline{24.8} \\ 
plants & 2.5 & 40.9 & 58.3 & 68.6 & 68.0 & \textbf{72.2} & \underline{71.6} \\ 
traffic-sign & 9.1 & 4.8 & 3.9 & 16.7 & \underline{22.2} & 18.2 & \textbf{29.1} \\ 
pole & 1.3 & 2.8 & 6.9 & 30.1 & 28.6 & \underline{31.5} & \textbf{37.8} \\ 
trashcan & 0.4 & 7.4 & 24.1 & 28.9 & 16.3 & \textbf{48.1} & \underline{46.3} \\ 
building & 37.1 & 57.5 & 66.1 & 75.1 & 73.1 & \underline{76.3} & \textbf{79.9} \\ 
cone/stone & 0.2 & 0.6 & 6.6 & \textbf{58.6} & 34.0 & 27.7 & \underline{34.5} \\ 
fence & 8.4 & 12.0 & 23.4 & 32.2 & 40.9 & \textbf{47.7} & \underline{47.3} \\
bike & 18.5 & 35.3 & 28.6 & 44.9 & 50.3 & 51.4 & \textbf{53.9} \\ 
ground & 72.1 & 71.3 & 73.5 & 76.3 & \underline{79.1} & \textbf{80.3} & 78.4 \\ \hline
mean-IoU & 12.9 & 26.9 & 30.9 & 43.2 & 46.4 & \underline{50.3} & \textbf{51.9} \\ \hline
\end{tabular}
}
\end{table}

%% file: tables/postprocess.tex
\begin{table*}[ht!]
\centering
\setlength{\tabcolsep}{3pt}
\caption{Comparison with different post-processing methods. Our MVP method is significantly better.}
\renewcommand{\arraystretch}{1.5}  
\resizebox{\textwidth}{!}{

        \begin{tabular}{c|l|ccccccccccccccccccc} \hline
         & \rotatebox{90}{{mean-IoU}}  & \rotatebox{90}{car}& \rotatebox{90}{bicycle}& \rotatebox{90}{motorcycle}& \rotatebox{90}{truck}& \rotatebox{90}{other-vehicle}& \rotatebox{90}{person}& \rotatebox{90}{bicyclist}& \rotatebox{90}{motorcyclist}& \rotatebox{90}{road}& \rotatebox{90}{parking}& \rotatebox{90}{sidewalk}& \rotatebox{90}{other-ground}& \rotatebox{90}{building}& \rotatebox{90}{fence}& \rotatebox{90}{vegetation}& \rotatebox{90}{trunk}& \rotatebox{90}{terrain}& \rotatebox{90}{pole}& \rotatebox{90}{traffic-sign} \\ 
 \hline\hline
        w/o MVP  & 60.4 & 85.8   & 44.0 &  61.5    & 80.3  & 53.0   & 68.7  & 70.2      & 0.91     & \textbf{94.8}   & 42.1    & 80.9     & 0.95         & 81.8    & 52.4  &  83.2       & 60.3  & 70.6    & 51.9 & 47.9         \\
        
        CRF~\cite{squeezeseg} &   \noimprove{58.2}{2.2} &  87.0  &  40.0 &  57.3  &  67.7  & 52.2 &  66.1  & 62.5 & 0.38  &  94.5 & \textbf{46.4} & \textbf{81.1}  &  0.66  & 81.7  &  53.6 &  81.4 &  60.9  & 66.3  &  49.0 & 47.6        \\
        
        PointRefine~\cite{motionseg3d}  &  \noimprove{59.2}{1.2} & 84.5 &  43.7 &  53.7 &  76.3 &  48.6 &  68.3 &  70.6 &  \underline{7.5} &  94.6 & 39.8 &  80.5 &  \textbf{11.8} &  81.4 &  50.7 &  83.8 &  59.4 &  \underline{72.2} &  51.1 &  46.1 \\
        
        NLA~\cite{fidnet}    &  \thirdimprove{64.4}{4.0} & 92.0 &  47.5  &  66.8  &  79.0 &  \underline{55.9}  &  76.2 &  \underline{85.7}  &   \textbf{12.4}  &  94.5 &   42.7  &   80.8   &   \underline{10.6}  &  87.3  &  54.6  &    85.9   &  66.0 &   \underline{72.2}  & 63.4 &  49.8        \\ 
        
        k-NN~\cite{rangenet}    & \secondimprove{64.5}{4.1} & \underline{91.4} & \underline{50.7}    & \underline{66.9}  & \underline{81.2}  &  54.9          & \underline{76.8}   & 85.1      & 0.96         & 94.5 & 41.6    & 80.9     & 0.95         & \underline{88.5}    & \underline{55.6}  & \underline{86.2}  & \underline{66.8}  & 71.5    & \underline{64.5} & \underline{50.2}   \\
        
        MVP (Ours)   & \bestimprove{66.5}{6.1} & \textbf{93.4} & \textbf{54.1}    & \textbf{70.2}       & \textbf{85.9}  & \textbf{59.8}         & \textbf{79.8}   & \textbf{88.0}      & 0.58         & \underline{94.7} & \underline{44.8}    & \textbf{81.1}     & 0.46         & \textbf{90.3}     & \textbf{66.6}  & \textbf{86.8}       & \textbf{69.5} & \textbf{72.7}    & \textbf{65.1} & \textbf{50.3}        \\\hline
        \end{tabular}
        }
\label{tab:post-process}
\end{table*}

%% file: sec/5_conclusion.tex
\section{CONCLUSION}

In this paper, 
we quantitatively and qualitatively analyzed the boundary blurriness, which is also called ``many-to-one'' problem, for range-image-based LiDAR segmentation, and introduced a novel solution named TFNet to tackle it.  
Our approach involves leveraging temporal information through the introduction of temporal fusion layers during the training process and a sequential max voting strategy during inference. 
The experiments on two benchmarks demonstrate the advantages of the proposed strategy. 
In particular, 
the incorporation of temporal data allows TFNet to maintain robust performance in environments with substantial occlusions, while still maintaining real-time performance.
Additionally, we conducted comprehensive ablation studies to validate the design, as well as the broader adaptability of the proposed post-processing to other neural network architectures.